%% file: main.tex
\PassOptionsToPackage{utf8}{inputenc}
\pdfoutput=1
\documentclass{bioinfo}
\usepackage{subfiles, tabularx,booktabs,ragged2e,lipsum,microtype,amsmath}
\usepackage{amssymb}
\usepackage{graphbox} 
\usepackage{makecell}

\usepackage[dvipsnames]{xcolor}
\renewcommand{\cite}{\citep} 
\copyrightyear{2015} \pubyear{2015}
\access{Advance Access Publication Date: Day Month Year}
\appnotes{Manuscript Category}

\begin{document}
\firstpage{1}

\subtitle{}

\title[Drug-target Counterfactual]{Counterfactual Explanation with Multi-Agent Reinforcement Learning for Drug Target Prediction}
\author[Nguyen \textit{et~al}.]{Tri Minh Nguyen\,$^{\text{\sfb 1,}*}$,  Thomas P Quinn \,$^{\text{\sfb 1}}$,  Thin Nguyen \,$^{\text{\sfb 1}}$ and Truyen Tran\,$^{\text{\sfb 1}}$}
\address{$^{\text{\sf 1}}$Applied Artificial Intelligence Institute, Deakin University, Victoria,
Australia\\}

\corresp{$^\ast$To whom correspondence should be addressed.}

\history{Received on XXXXX; revised on XXXXX; accepted on XXXXX}

\editor{Associate Editor: XXXXXXX}

\subfile{abstract.tex}

\maketitle

\section{Introduction}
\input{intro.tex}
\section{Related works}
\input{relatedworks.tex}

\section{Methods and materials}
\subsection{Problem definition}
\label{sec:problem}
\input{method_problemstatement}

\subsection{Counterfactual generation with RL}
\label{sec:SingleAC}
\input{method_CF_RL}
\subsection{Drug-protein pair counterfactual generation with MARL}
\input{method_overview.tex}

\subsubsection{Available drug actions}
\label{subsec:Drug-gen}
\input{method_drug.tex}

\subsubsection{Available protein actions} \label{sec:protgen}
\input{method_protein.tex}

\subsubsection{Multi-agent actor-attention-critic for counterfactual generation}
\label{sec:maac}
\input{method_MAAC.tex}

\input{exp.tex}

\section{Results and Discussion}
\input{result_discuss}

\section{Conclusion}
\input{conclusion.tex}

\bibliographystyle{natbib}
\bibliography{tri,thom}









\end{document}

%% file: abstract.tex
\abstract{\textbf{Motivation:}
Many high-performance DTA models have been proposed, but they are
mostly black-box and thus lack human interpretability.
Explainable AI (XAI) can make DTA models more trustworthy,
and can also enable scientists to distill biological knowledge from the models. 
Counterfactual explanation is one popular approach to explaining the
behaviour of a deep neural network, which works by systematically
answering the question ``How would the model output change if the
inputs were changed in this way?''.
Most counterfactual explanation methods
only operate on single input data. 
It remains an open problem how to extend counterfactual-based XAI methods to DTA models, which have two inputs, one for drug and one for target, that also happen to be discrete in nature.
\\
\textbf{Methods} We propose a multi-agent reinforcement learning
framework, Multi-Agent Counterfactual Drug-target binding Affinity
(MACDA), to generate counterfactual explanations for the drug-protein
complex. Our proposed framework provides human-interpretable counterfactual
instances while optimizing both the input drug and target for counterfactual
generation at the same time.
\\
\textbf{Results:} We benchmark the proposed MACDA framework using the Davis dataset and find that our framework produces more parsimonious explanations with no loss in explanation validity, as measured by encoding similarity and QED. We then present a case study involving ABL1 and Nilotinib to demonstrate how MACDA can explain the behaviour of a DTA model in the underlying substructure interaction between inputs in its prediction, revealing mechanisms that align with prior domain knowledge.\\
 \textbf{Availability:} The Python implementation is available at
https://github.com/ngminhtri0394/MACDA\\
 \textbf{Contact:} \href{http://minhtri@deakin.edu.au}{minhtri@deakin.edu.au}\\
}

%% file: intro.tex
Drug-target binding affinity (DTA) prediction is an important step
in drug discovery and drug repurposing \cite{Thafar2019ComparisonAffinities}.
Many high-performance DTA models have been proposed, but they are
mostly black-box and thus lack human interpretability \cite{Ozturk2018DeepDTA:Prediction,Ozturk2019WideDTA:Affinity,Torng2019GraphInteractions,Zheng2020PredictingSystem,Jiang2020DrugtargetMaps,Nguyen2020GraphDTA:Networks,Tri2020GEFA:Prediction}.
This lack of interpretability makes it difficult to use deep learning
models to \textit{distill knowledge} about how drugs bind to their
targets. Explainable AI methods explain how a model works, and therefore
facilitate human knowledge distillation. When applied to DTA prediction,
model explanations can produce insights that feedback into the
research pipeline and improve our understanding of drug-molecule binding. 

 

\begin{figure}
\begin{tabular}{|P{2cm}|c|c|c|}
\hline
Instance & Drug & Protein & $\Delta$ Affinity \\
\hline
\begin{tabular}{p{2cm}p{2cm}}Original drug \\ Original protein\end{tabular} &
\includegraphics[width=2.5cm,align=c]{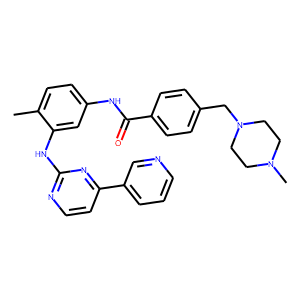}&...PFWKYY...& 0 \\
\hline
\begin{tabular}{p{2cm}p{2cm}}Counterfactual drug \\ Original protein\end{tabular}
&\includegraphics[width=2.5cm,align=c]{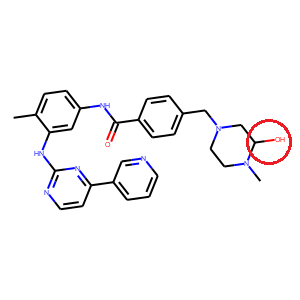}&...PFWKYY...&High\\
\hline
\begin{tabular}{p{2cm}p{2cm}}Original drug \\ Counterfactual protein\end{tabular}&\includegraphics[width=2.5cm,align=c]{img/Imatinib_org.png}&...P\textcolor{red}{\textbf{A}}WKYY...&High\\
\hline
\begin{tabular}{p{2cm}p{2cm}}Counterfactual drug \\ Counterfactual protein\end{tabular}&\includegraphics[width=2.5cm,align=c]{img/Imatinib_cf3_highlight.png}&...P\textcolor{red}{\textbf{A}}WKYY...&Low\\
\hline
\end{tabular}
\caption{A summary of the \textit{unstable environment problem} in drug-target binding affinity counterfactual generation. If the drug or target is fixed, the $\Delta$ affinity is high. However, the $\Delta$ affinity between counterfactual instance of drug and protein may be low as they are not optimized for new drug/protein.}
\label{fig:difficult_MADTA}
\end{figure}

Counterfactual explanation is one popular approach to explaining the
behaviour of a deep neural network, which works by systematically
answering the question ``How would the model output change if the
inputs were changed in this way?''. Research into this type of explainable AI has mostly focused on image data and tabular data \cite{Vermeire2020ExplainableCounterfactual,Mothilal2020ExplainingExplanations,Dhurandhar2018ExplanationsNegatives,Cheng2020DECE:Models}.
In the context of the drug-target affinity, counterfactual explanations
are not widely used. Instead, explanation methods rely on feature
attribution scores and gradients \cite{Preuer2019InterpretableDiscovery,Pope2019ExplainabilityNetworks,McCloskey2019UsingChemistry},
which may fail to capture high-order interactions between features
\cite{tsang_neural_2018}. It remains an open problem of how to produce
counter-factual explanations for drug-target affinity models.

There are two key challenges in extending counterfactual explanations
to drug-target affinity models. First, the inputs to a DTA model,
represented most often as sequences or graphs, are discrete not continuous.
Therefore, gradient-based counterfactual generation methods, which
operate on the continuous data, cannot be applied. Meanwhile, gradient-free
combinatorial methods like \textit{in silico mutagenesis} \cite{zhou_predicting_2015}
are computationally expensive to run. Second, DTA models have two
distinct inputs, the drug and the target. Changes to the drug
molecule and protein can both influence the binding affinity, either separately
or jointly. Generating counterfactuals for drug and target separately
may not lead to an optimal solution (see Fig.~\ref{fig:difficult_MADTA}). On top of that, the substructure interactions between drug functional groups and protein residues are crucial DTA model prediction. However, it is difficult to disentangle and interpret the effect of substructure interaction from the DTA model prediction. 

We propose \textbf{M}ulti-\textbf{A}gent \textbf{C}ounterfactual \textbf{D}rug-target
binding \textbf{A}ffinity (\textbf{MACDA}) framework, which uses multi-agent
reinforcement learning (\textbf{MARL}) to solve both challenges. First, MARL
solves the challenge of discrete inputs because MARL is naturally designed
for \textit{discrete actions}. These actions can serve as a fundamental unit of human-interpretation. In the DTA problem, the adding or removing of
bonds and atoms on a drug can be thought of as actions (as can the adding or removing of motifs and residues on a protein).
Second, MARL solves the
challenge of multiple inputs because the drug and the protein can
be represented as different action space inputs within a single model. Third, MACDA isolates the effect of substructure interaction of drug and protein in the model prediction. This allows domain experts to validate the binding mechanism the DTA model used in its prediction.

In contrast to previous work that used reinforcement learning (RL) to facilitate drug synthesis \cite{Zhou2019OptimizationLearning}, we use RL here to solve a more general problem of explaining the behavior of black-box DTA model in order to understand the biology of drug-protein complex.
To this end, we propose the multi-agent RL framework,
MACDA, for explaining deep models of drug-target binding affinity. We evaluate the proposed framework on the publicly available
Davis dataset, where we observe a stable learning process and counterfactual
explanations that agree with the state-of-the-art method in molecule
counterfactual generation.

%% file: relatedworks.tex
\subsection{Drug-target affinity models}

Drug-target binding affinity, measured by a disassociation constant
$K_{d}$, indicates the strength of the binding force between the target
protein and its ligand (drug or inhibitor). Drug-target binding affinity
prediction methods can be categorized into two main approaches: structural
approach and non-structural approach \cite{Thafar2019ComparisonAffinities}.
The structural approach \cite{Meng1992AutomatedEvaluation,Jorgensen1983ComparisonWater,Pullman2013IntermolecularForces,Raha2007TheDesign}
uses the 3D information of the protein structure and ligand to run
a drug-target interaction simulation. The non-structural
approach \cite{Nguyen2020GraphDTA:Networks,Ozturk2018DeepDTA:Prediction,Ozturk2019WideDTA:Affinity,Tri2020GEFA:Prediction}
uses other information such as protein sequence, atom valence, hydrophobic,
and others to 
from existing databases to 
predict the binding affinity. 
In MACDA, we take the non-structural approach.

\subsection{Explaining deep neural networks for DTA}

Recently, high accuracy DTA models are often based on deep neural networks, which are mostly black-box. This poses a great question of how to explain the behaviours of the deep models, and distill the explicit knowledge from them. 
There are two general approaches relevant to the explanation of drug-target affinity models: feature attribution and graph-based
methods.

Feature attribution measures the relevance score of the input feature
with respect to the predicted affinity score $y$, either using the gradient
\cite{Preuer2019InterpretableDiscovery,Pope2019ExplainabilityNetworks}
or a surrogate model \cite{Rodriguez-Perez2020InterpretationValues}. Gradient-based
methods take advantage of the derivative of the output with respect
to the input. \citet{McCloskey2019UsingChemistry} use integrated gradients
on graph convolution model trained on the molecular binding synthesis
dataset to analyze the binding mechanism. However, gradient-based
methods could be misleading or prone to gradient saturation \cite{Ying2019GNNExplainer:Networks}.
Surrogate-based methods generate a surrogate explanatory model $g$
which is interpretable (linear or decision tree) and can approximate
the original function $f$. 

Graph-based methods are suitable for DTA because the structure of
the drug molecule can be represented naturally with the graph structure.
The graph can be explained by subsets of edges and node features which
are important for model $f$ prediction of class $c$. For example,
GNNExplainer \cite{Ying2019GNNExplainer:Networks} finds a subgraph $G'$
of input graph $G$, and subfeature $X'$ of input feature $X$ which
maximizes the mutual information between $f(G,X)$ and $f(G',X')$,
but is argued to not generalize well \cite{Numeroso2020ExplainingCounterfactuals}.
Attention-based graph neural network \cite{Velickovic2018GraphNetworks,Shang2018EdgeNetworksb,Ryu2018DeeplyNetwork}
is a mechanism that can facilitate explanation such as the influence
of substructure to the solubility property \cite{Shang2018EdgeNetworksb},
visualizing the importance of neighbor nodes via an attention score.
In MACDA, we use a graph-based method.

\subsection{Reinforcement learning: Single and Multi-agent}

Reinforcement learning (RL) is the process of agent learning to find
the optimal action for situations that maximizes the long-term rewards
\cite{Sutton2018ReinforcementIntroduction}. For the single agent case, an agent
interacts with the environment. At each time step $t$, the agent
observes environment state $s_{t}\in S$ and chooses an action $a_{t}\in A$
using its policy $\pi(a_{t}|s_{t})$. By completing the action, the
agent receives a reward $r_{t}$ and changes the environment to the
next state $s_{t+1}$.

There are two main approaches to RL: value-based and policy-based.
The value-based methods estimate the value function for each state, 
where the action is chosen based
on the action value. 
The policy-based methods, on the other hand, optimize the agent's
policy as a function $\pi(a|s,\theta)$ where $\theta$ is the parameter. The two
methods can be combined, e.g., both value function and the policy
are estimated, as in the celebrated actor-critic methods \cite{Konda1999Actor-CriticTypeProcesses,Morimura2009AAlgorithm}.
In MACDA, we use an actor-critic method.

Multi-agent reinforcement learning (MARL) is the generalization from
a single agent to multiple agents that share the same environment.
Each agent interacts with the environment and with other agents. The
challenge of multi-agent RL is finding the optimal policy for each
agent with respect to not only the environment but also to the other agent's
policy. Many approaches solving the multi-agent setting have been
proposed, ranging from cooperative communication \cite{Tan1993Multi-AgentAgents,Fischer2004HierarchicalCoordination}
to competitive environment \cite{Littman1994MarkovLearning,Perez-Liebana2019TheCompetition}.
In MACDA, we use cooperative communication.

\subsection{Using reinforcement learning for explanations}

Instead of assigning a relevance score to the input features of
the model, counterfactual explanation finds the simplest perturbed
instance with a maximal difference in model prediction outcome \cite{Wachter2017CounterfactualGDPR}.
The motivation here is that if a small change in part of the input causes a big change in the output, then that part of the input is important.

Reinforcement learning has been used previously to generate counter-factual explanations
\cite{Hendricks2016GeneratingExplanations,Li2016UnderstandingErasure}. \citet{Numeroso2020ExplainingCounterfactuals} generate a counterfactual
explanation for a molecule using MEG framework, a multi-objective
reinforcement learning, to maximize the prediction model output change
while maintaining the similarity between original and counterfactual molecule instances.
We use the multi-agent version of the MEG framework as the baseline
for our experiments. Compared to multi-agent MEG framework, our proposed
MACDA framework uses multi-agent actor-critic approach in which each modification to drug and protein are considered simultaneously.

%% file: method_problemstatement.tex
A counterfactual explanation is given as a hypothetical statement "If X had been X', Y would have been Y'" \cite{Pearl2016CausalPrimer,Goyal2019CounterfactualExplanations}. The counterfactual explanation $X'$ is an instance which resembles $X$ while leading to a substantially different outcome. To give a counterfactual example of the instance $X$, we formulate as an optimization problem:
\begin{equation}\label{eq:op_singlecf}
    arg \max_{X} \lambda y_{loss}(Y-Y') + sim(X, X')
\end{equation}
where $sim(X,X')$ is the similarity between the original instance and counterfactual instance, while $\lambda$ balances the prediction distance $y_{loss}$ and the similarity.

We extend the hypothetical statement for drug-target pair $(D,P)$ interaction in DTA model $\mathcal{F}$ as "If drug $D$ and protein $P$ interaction had been $D'$ and $P'$ interaction, the predicted affinity $\mathcal{F}(D,P)$ would have been $\mathcal{F}(D',P')$".

Given a drug-target pair $(D,P)$ with the predicted binding affinity $\mathcal{F}(D,P)$ in DTA model $\mathcal{F}$, the task of finding the counterfactual interaction pair $(D',P')$ in $\mathcal{F}$ is the optimizing problem.
\begin{align}
\begin{split}
\label{eq:op_DTA}
arg \max_{D',P'} & \lambda_1 y_{loss}(\mathcal{F}(D,P)-\mathcal{F}(D',P'))\\
 & + \lambda_2 sim(D, D') + \lambda_3 sim(P, P')\\
\end{split}
\end{align}
where $sim(D, D')$ and $sim(P, P')$ measure the similarity between the input instance and counterfactual instance of drug and protein, while $\lambda_1$, $\lambda_2$, and $\lambda_3$ are the weight coefficients.

%% file: method_CF_RL.tex
In this section, we briefly describe the process of generating the counterfactual instance using RL. We choose RL to solve the optimizing the function in Eq. \ref{eq:op_DTA} as RL can operate on discrete action space (e.g., to change an atom or amino acid residue). Specifically, we employ Actor-Critic RL \cite{Sutton2018ReinforcementIntroduction}. Unlike Q-learning which tries to evaluate the Q-value of all actions and choose an action with greedy algorithm, ACRL learns to choose the action robustly with policy network $\pi$. In ACRL, a Q-value network assists the policy network by updating the policy network parameter $\theta$. The intuition behind the policy network $\pi$ and Q-value network $Q$ in ACRL is that the policy network is the actor, learning a robust strategy, and the Q-value network is the critic, correcting the actor strategy.

Our ACRL framework consists of two main components: \textbf{(a)} the agent and \textbf{(b)} the simulated environment with the model $\mathcal{F}$ as its core. The agent choose an \textbf{action} and \textbf{interacts with the environment}. By \textbf{action} we mean that the agent chooses a modified version of the input drug or protein. By \textbf{interacts with the environment}, we mean that the model $\mathcal{F}$ receives a modified version of drug and protein, then suggests their simulated binding affinity.

The environment has the model $\mathcal{F}$ as its core to calculate the binding affinity between drug and protein generated by agents. After calculating the affinity, the environment will return the \textbf{state} of drug and protein and the \textbf{reward} (see Fig. \ref{fig:mul_AC}). The returned state is a Morgan fingerprint for drug and a protein sequence for target. The reward goal is to optimize the counterfactual function Eq. \ref{eq:op_DTA}. We will further describe the reward function in Sec. \ref{sec:maac}. 

The agent handles the counterfactual example generation based on the environment state and reward. In our case, the agent generates a modified version of the input drug and protein state. The ACRL agent has two mains components: \textbf{(a)} policy network with parameter $\theta$ and \textbf{(b)} Q-value network with parameters $\psi$.

The Q-value network $Q_{\psi}(s,a)$ plays a role as critic, updated using Q loss (see Fig. \ref{fig:mul_AC}). Given an action-state $(s,a)$ with reward $R_t$ at time step $t$ and action-state $(s',a')$ at time step $t'=t+1$, the parameter $\psi$ is updated with the Q loss function:
\begin{equation}
    \mathcal{L}_Q = R_t + \gamma Q_{\psi}(s',a') - Q_{\psi}(s,a)
\end{equation}
where $\gamma$ is the discount factor.
The policy network $\pi_{\theta}$ decides which action to take. A policy is the probability of each action being selected. The actor updates the $\theta$ parameter of the policy network with the suggestion of critic:
\begin{equation}
    \theta \longleftarrow \theta + \alpha Q_{\psi}(s,a)\nabla_{\theta}\log\pi_{\theta}(a|s)
\end{equation}
where $\alpha$ is the learning rate.

%% file: method_overview.tex

We now describe our MARL framework which is the extension from the ACRL framework (in Sec. \ref{sec:SingleAC}) for generating drug-target counterfactuals. MARL is particularly suitable because it works naturally on multiple discrete action spaces. In our setting, the two action spaces correspond to the discrete modifications in the molecule space and protein space, respectively.
In particular, we will employ a MARL framework known as Multi-agent Actor-Attention-Critic (MAAC) \cite{Iqbal2019Actor-Attention-CriticLearning}. This framework is flexible, easy to train, and is natural for exploring the joint space of protein-drug complex. MAAC allows separated policies for drug and protein, but with common critics and rewards. It uses the attention mechanism to dynamically select relevant information shared by the other agent, a procedure that resembles the selective binding mechanism often found in the protein-drug complex \cite{Tri2020GEFA:Prediction}. 

\begin{figure}
\centering{}\includegraphics[width=0.5\textwidth]{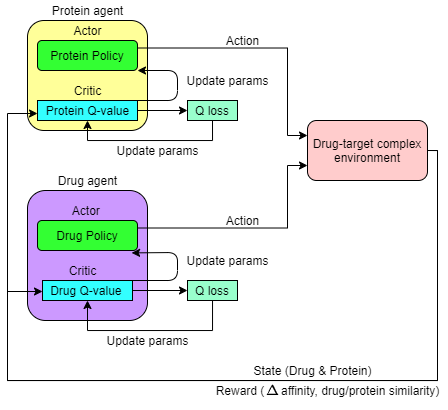} \caption{The overview of MARL framework for drug-target counterfactual generation.
Two agents, protein and drug, have their own actor and critic. The
actor chooses the action using its policy. Two agents take actions
and the drug-target environment returns (a) the new state
of drug and protein, and (b) the reward, which is a sum
of the prediction change and the protein/drug similarities. Then, the critic calculates its Q value using
the reward value. The Q value is used to update the critic network.
\label{fig:mul_AC} }
\end{figure}

The framework is illustrated in Fig.~\ref{fig:mul_AC}. There are
two agents: one generates counterfactuals for the protein,
and the other for the drug. The two agents work in tandem to produce joint counterfactuals for the protein-drug complex. Each agent
has its own $Q$-value function. Two agents communicate through the reward function described in Eq. \ref{eq:reward} and through the $Q$-value function (Eq. \ref{eq:MAQvalue}).

In what follows, we describe the framework components. In particular,
the action space for drug and protein are provided in Sec.~\ref{subsec:Drug-gen}
and Sec.~\ref{sec:protgen}. The overall reward is
presented in Sec.~\ref{sec:maac}.

%% file: method_drug.tex

We adopt the drug molecule generation strategy in Mol-DQN \cite{Zhou2019OptimizationLearning}.
There are three action categories: (a) \textbf{Add atom}: Given an
admissible set of atoms $\mathcal{E}=\{Atom_{1},..,Atom_{N}\}$, one
atom $Atom_{i}$ is inserted into the drug molecule at a time. Then
a bond is formed between the newly added atom and a position satisfying
the valence constraint. Therefore, given $n_{p}$ positions satisfying
the constraint, there are $n_{p}$ instances generated. (b)\textbf{
Add bond}: One or more bond is added up to triple bond between two
atoms with free valence. (c)\textbf{ Remove bond}: One or more bond
is removed from the existing bond. If there is no bond between two
atoms after removal, then the disconnected atom is removed. 

All possible actions are generated and used as action space in the
reinforcement learning framework described in Sec.~\ref{sec:maac}.

%% file: method_protein.tex
For proteins, there is only one action category: (a) \textbf{Replace amino acid resiue with alanine}.
Alanine is widely used to determine the
contribution of protein residues in the protein function or drug-protein
binding \cite{Gray2017AnalysisSubstitutions}. Alanine is chosen as its size is
not too large which avoids steric hindrance. In addition, the methyl
function group allows it to mimic the secondary structure of the residues
it replaces (\cite{Gray2017AnalysisSubstitutions}). This process is known as alanine scanning \cite{Gray2017AnalysisSubstitutions}. For the protein sequence $P={r_{i}},i\leq l$
where $l$ is protein sequence length, a single residue $r_{i}$ is
replaced with alanine to create a single point alanine mutation. All
possible single point mutations are generated and used as the action space.

%% file: method_MAAC.tex

The idea of multi-agent actor-critic is that the Q-value function
of agent $i$ is calculated based on the observation of other agents
$s=(s_{1},...,s_{N})$: 
\begin{equation}
Q_{i}^{\theta}=f_{i}\left(g_{i}(s_{i},a_{i}),x_{i}\right)\label{eq:MAQvalue}
\end{equation}
where $f_{i}$ and $g_{i}$ are multi-layer perceptron, $x_{i}$ is
the weighted sum of other agents value: 
\begin{equation}
x_{i}=\sum_{j\neq i}\alpha_{ij}\sigma\left(Vg_{j}(s_{j},a_{j})\right)\label{eq:AgentQvalue}
\end{equation}
where $V$ is the linear transformation, $\sigma$ is Leaky ReLU,
and $\alpha_{ij}$ is the attention score computed as in \cite{Vaswani2017AttentionNeed}
by taking $g_{i},g_{j}$ as the inputs. Learning in this actor-critic
framework then proceeds for each agent using the framework described
in Sec.~\ref{sec:SingleAC}.

In our context of protein-drug counterfactual generations, this boils
down to using state-action function of drug agent to influence the
Q-function of the protein agent and vice versa. In particularly, the Eq. \ref{eq:MAQvalue} and Eq. \ref{eq:AgentQvalue} for drug agent become:

\begin{equation}
    Q_{d}^{\theta}=f_{d}\left(g_{d}(s_{d},a_{d}),x_{d}\right)
\end{equation}
where
\begin{equation}
x_{d}=\sigma\left(Vg_{p}(s_{p},a_{p})\right)
\end{equation}
The Q-value function for protein is calculated in the similar manner.

%


\textbf{Multi-objective reward function}: Our objective is to find the counterfactual
satisfying two constraints: (1) to maximize the change in the predicted
binding affinity, and (2) to maximize the similarity between original instance and counterfactual instance. 

The $\Delta$ affinity between the predicted value of counterfactual drug-protein $(D',P')$ and the predicted value of the original instance $(D,P)$, $\Delta\mathcal{F}(D',P')$, is calculated as:

\begin{equation} \label{eq:totaldeltalaff}
    \Delta\mathcal{F}(D',P') = y_{loss}(\mathcal{F}(P',D')-\mathcal{F}(P,D))
\end{equation}
Here, we choose $y_{loss}$ to be absolute error. 

Our target is identifying the joint importance of drug sub-structures and protein residues using the joint counterfactual. In other words, we identify which parts of the drug and protein interacting in the DTA model. However, the $\Delta\mathcal{F}(D',P')$ only considers the consequences of the total drug-target counterfactual. It does not isolate the effect of the joint counterfactual. The joint drug-target counterfactual $\Delta\mathcal{F}$ affinity change is given as:

\begin{align}
\begin{split}
\label{eq:deltajoint}
\Delta_{joint}\mathcal{F}(D',P') & = \Delta\mathcal{F}(D',P') - \Delta\mathcal{F}(D',P) - \Delta\mathcal{F}(D,P')\\
\end{split}
\end{align}
where $\Delta\mathcal{F}(D,P')$ and $\Delta\mathcal{F}(D',P)$ are counterfactual target - fixed drug $\Delta\mathcal{F}$ affinity, and vice versa. Simply speaking, the joint drug-target counterfactual $\Delta\mathcal{F}$ affinity shows contribution of perturbing both drug and protein jointly. Interacting drug-substructure and protein residue should change the affinity prediction more than the sum of the individual perturbations. A high $\Delta_{joint}\mathcal{F}(D',P')$ shows that the DTA model factors the drug-substructure and protein residue interaction in its prediction.

As our goal is identifying the interacting drug sub-structure residue pair, we add the sign function to give negative reward when the generated $(D', P')$ pair increases the affinity:

\begin{align}
\begin{split}
\label{eq:deltajoint_sign}
\Delta_{sjoint}\mathcal{F}(D',P') & = -\text{sign}\left(\mathcal{F}(P',D')-\mathcal{F}(P,D)\right) \times \Delta\mathcal{F}(D',P')\\
 & \quad\quad- \Delta\mathcal{F}(D',P) - \Delta\mathcal{F}(D,P')
\end{split}
\end{align}

 The similarity is the cosine similarity:
\begin{equation}
\text{d}(\mathcal{F}_{e}(x),\mathcal{F}_{e}(y))=\frac{\mathcal{F}_{e}(x)\cdot \mathcal{F}_{e}(y)}{\|\mathcal{F}_{e}(x)\|\|\mathcal{F}_{e}(y)\|}.\label{eq:sim}
\end{equation}
where $\mathcal{F}_{e}$ is the encoded representation of drug or protein in the DTA model. In case the encoded representation is not available, the Tanimoto similarity between two molecule fingerprints and protein sequence similarity can be used as the alternative similarity.

Then the reward function is defined as: 
\begin{align}
\begin{split}
\label{eq:reward}
R(s) & =\alpha_{r}\Delta_{sjoint}\mathcal{F}(D',P')+\alpha_{p}\text{sim}(\mathcal{F}_{e}(P),\mathcal{F}_{e}(P'))\\
 & \quad\quad+\alpha_{d}\text{sim}(\mathcal{F}_{e}(D),\mathcal{F}_{e}(D'))
\end{split}
\end{align}
where $P'$ and $D'$ are the counterfactual instance of drug $D$
and protein $P$, $\alpha_{r}$, $\alpha_{d}$. and $\alpha_{p}$ are coefficients that balance between the predicted affinity change and the similarity. 
We add both protein and drug molecule similarity terms to the reward. First, both similarity terms help the model to generate molecules and sequences with minimal change, satisfying one of counterfactual constraints. Second, it works as a communication between two agents where the drug agent searches for a molecule that does not require significant change in protein and vice versa. The first term, $\Delta_{sjoint}\mathcal{F}(D',P')$, is to meet the model output change constraint. The two similarity terms encourage the similarity between original instance and counterfactual instance.

%% file: exp.tex
\subsection{Dataset}

We evaluate our method MACDA on the Davis dataset \cite{Davis2011ComprehensiveSelectivity},
which contains the drug-target binding affinity of 442 target proteins
and 72 drugs. We use $pK_{D}$ (log kinase dissociation constant) to measure the binding affinity between the target protein and
the drug molecule, similar to \cite{Ozturk2018DeepDTA:Prediction}. The drug-target pairs between Tyrosine-protein
kinase ABL1 (Human) and 50 drugs in the training set are chosen to
generate counterfactual instances. We choose ABL1 as the crystallized complex of ABL1 with various drugs are available for evaluation.  

\subsection{Baselines}
We compare MACDA with 2 baseline methods.
First, for \textbf{Joint-List}, we choose the top ten drug and protein counterfactual instances having highest $\Delta$ affinity and similarity separately (i.e., top 10 for drugs and top 10 for proteins). Then, the two lists are joined to form drug-protein counterfactual instances. Second, for \textbf{MA-MEG}, we extend the molecule counterfactual generation MEG framework \cite{Numeroso2020ExplainingCounterfactuals}
to the drug-target counterfactual generation task. As the MEG framework
only has a single agent handling the optimization for drug molecule,
we add another agent handling the protein sequence optimization. The
protein agent has the action space described in Sec.~\ref{sec:protgen}.
The protein agent calculates and updates its Q-function in the same
manner as the drug agent. Two agents work independently to optimize
the common reward function (see Eq.~(\ref{eq:reward})).

\subsection{Implementation detail}

The MACDA framework is implemented in Python using Pytorch. GraphDTA-GCNNet
\cite{Nguyen2020GraphDTA:Networks} is used as a drug-target binding
affinity prediction model because of its simplicity and high performance.
GraphDTA-GCNNet receives the drug molecule graph and protein sequence
as the inputs. The drug molecule observation in MACDA framework is
the drug fingerprint. The protein observation in MACDA framework is
the alphabet sequence encoded to integer sequence. The protein sequence
length is fixed at 1000 residues. To follow the similarity constraint and alanine scanning procedure \cite{Numeroso2021MEG:Networks}, we set the original drug molecule and protein sequence as the starting point and set the episode length to 1. The balance coefficients $\alpha_{r} = 1.0$, $\alpha_{d} = 0.05$. and $\alpha_{p} = 0.01$ in Eq. \ref{eq:reward} are chosen based on our experience.

For each drug-target instance, the top ten counterfactual instances
with the highest reward are chosen. The hyperparameters in the experiment
are shown in Table~\ref{tab:hyperparam}. The hyperparameters are chosen based on our experience. 

\begin{table}[h]
\centering{}\caption{The hyper-parameters used in the experiments.\label{tab:hyperparam}}
\begin{tabular}{p{7cm}p{1cm}}
\toprule 
Hyper parameters  & Value\tabularnewline
\midrule 
$\gamma$  & 0.99\tabularnewline
Batch size  & 1024 \tabularnewline
Policy learning rate  & 0.001 \tabularnewline
Critic learning rate  & 0.001\tabularnewline
Number of episode  & 10000\tabularnewline
\bottomrule
\end{tabular} 
\end{table}

\subsection{Evaluation metrics}

Two methods are evaluated using four metrics: average
drug encoding similarity, average protein encoding similarity, average $\Delta_{joint}$ between the predicted binding affinity of
original and counterfactual instance defined in Eq. \ref{eq:deltajoint}, and drug-likeness (QED) \cite{Bickerton2012QuantifyingDrugs}. The similarity
score is defined in Eq.~(\ref{eq:sim}). The protein, drug similarity, and the $\Delta$ affinity evaluate how good the generated counterfactual is. These two metrics follows the counterfactual definition in Sec. \ref{sec:problem}. The $\Delta$ affinity metric makes sure the generated counterfactual has substantial change in the affinity. The drug and protein encoding similarity estimates how much the generated drug and protein resemble the original instance. The QED assesses the validity of the drug counterfactual
instance \cite{Bickerton2012QuantifyingDrugs}. Because we use alanine scanning as the protein single point mutation, the protein sequence does not change significantly. Therefore, the protein sequence validation is not necessary.

%% file: result_discuss.tex
\subsection{MACDA produces highly parsimonious explanations}

Table~\ref{tab:result} shows a comparison of our proposed MACDA framework with baseline methods, where our method exhibits state-of-the-art average $\Delta_joint$ affinity, drug similarity, protein similarity, and QED. The first baseline, \textbf{Joint-List}, simply chooses the top drug and protein counterfactuals, then joins them together. As such, it cannot find an interacting counterfactual pair. Thus, the average $\Delta_{joint}$ of the first baseline is negative. The second baseline, \textbf{MA-MEG} does better, having an average $\Delta_{joint}$ that is above zero, but still underperforms MACDA.

QED measures the drug-likeness of the molecules, providing an estimate of the validity of a generated drug.
In a variant to MACDA, called \textbf{MACDA-QED}, we incorporate the QED into the reward function to increase the validity of the generated drug counterfactual. As a result, MACDA-QED increases the QED by $13.1\%$ compared to the QED of the original data. As the QED is higher, the drug similarity is also slightly higher. However, as a trade off, the average $\Delta_{joint}$ is lower since the QED imposes another constraint on the generated drug distribution. Therefore, the counterfactual drug distribution is closer to the original drug distribution.

The average drug similarity and protein similarity columns show us that the MACDA counterfactuals are very similar to the original drug/protein input. Yet, the changes have a big impact on $\Delta_{joint}$.
We consider MACDA explanations to be more parsimonious because small changes in the input can produce big changes in joint affinity prediction.

\begin{table}[h]
\caption{The average $\Delta_{joint}$, drug encoding similarity, target
encoding similarity, and QED. MACDA is highly parsimonious in that it can find small changes to the input that produce big changes to the joint predicted affinity.\label{tab:result}}
\resizebox{\columnwidth}{!}{{%
\begin{tabular}{p{2cm}ccccc}
\toprule 
Method  & Avg. $\Delta_{joint}$ $\uparrow$  & Avg. Drug Sim.$\uparrow$  & Avg. Protein Sim. $\uparrow$  & QED $\uparrow$\tabularnewline
\midrule 
\textbf{Original drug/protein} & 0 & 1 & 1 & 0.4366 \tabularnewline
\textbf{Joint-List baseline} (Ours) & -0.0085 & 0.9208 & 0.9992 & 0.4051 \tabularnewline
\textbf{MA-MEG*} \cite{Numeroso2020ExplainingCounterfactuals}  & 0.0178  & 0.9274  & \textbf{0.9993}  & 0.4086 \tabularnewline
\textbf{MACDA} (Ours)  & \textbf{0.0254}  & 0.9209  & \textbf{0.9993}  & 0.4056 \tabularnewline
\textbf{MACDA-QED} (Ours)  & 0.0224  & \textbf{0.9481}  & \textbf{0.9993}  & \textbf{0.4586} \tabularnewline
\midrule
\multicolumn{5}{l}{(*) Our extended version of single-agent MEG framework}\tabularnewline
\bottomrule
\end{tabular}}{ } } 
\end{table}

\subsection{MACDA explains DTA model binding site}

\begin{figure}[h]
\centering{}\includegraphics[width=0.5\textwidth]{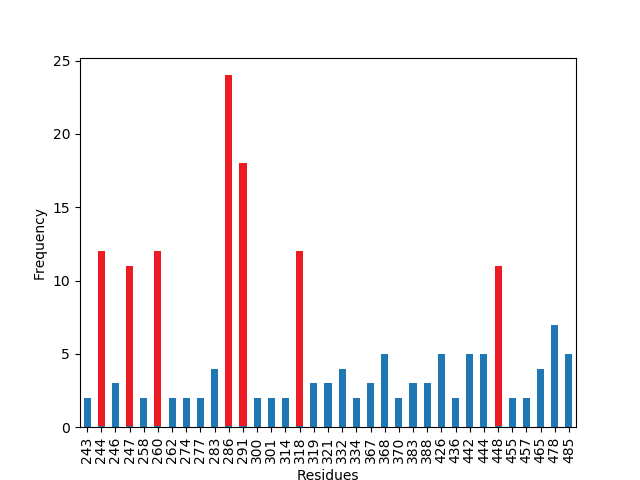}
\caption{The mutation point distribution in the kinase domain of ABL1 over 500 counterfactual instances
of 50 ABL1-drugs pairs. The top occurrences of alanine replacements are highlighted in red: residues MET.244, LYS.247, VAL.260, GLU.286, LYS.291, and VAL.448 cause the highest change in binding affinity. \label{fig:prot_mutation_pos}}
\end{figure}
\begin{figure}[h]
\centering{}\includegraphics[width=0.45\textwidth]{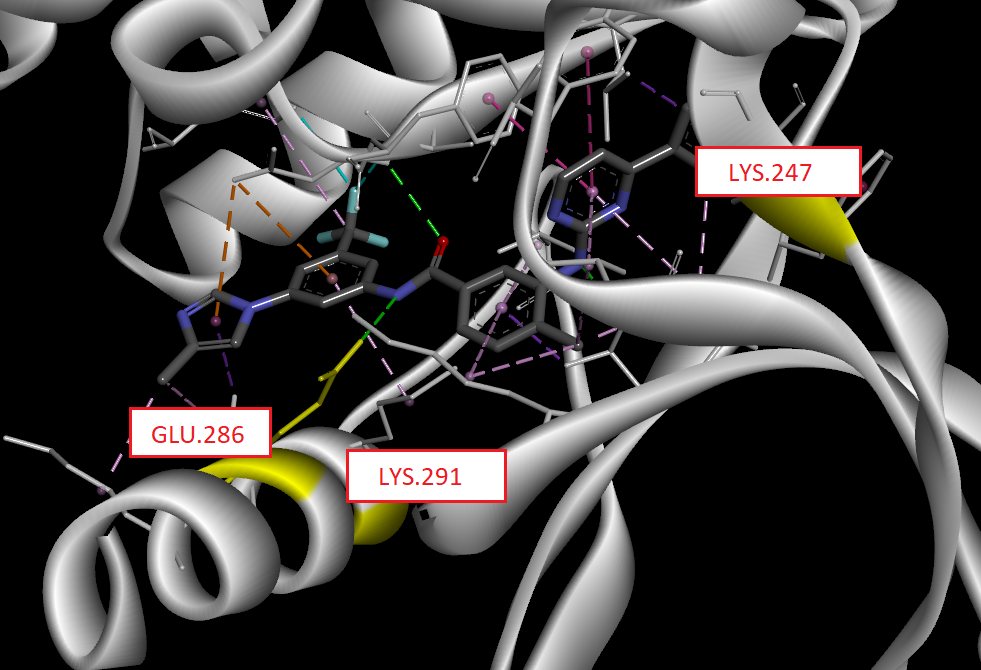}
\caption{Visualization of residue LYS.247, GLU.286, and LYS.291 of protein ABL1 in the ABL1-Nilotinib (PDB
3CS9). Note the proximity between Nilotinib and residues LYS.247, GLU.291. Another important residue, GLU.286, is the binding site of ABL1-Nilotinib. Figure best viewed in color. \label{fig:ABL1_Nilotinib}}
\end{figure}
\begin{figure}[h]
\centering{}\includegraphics[width=0.45\textwidth]{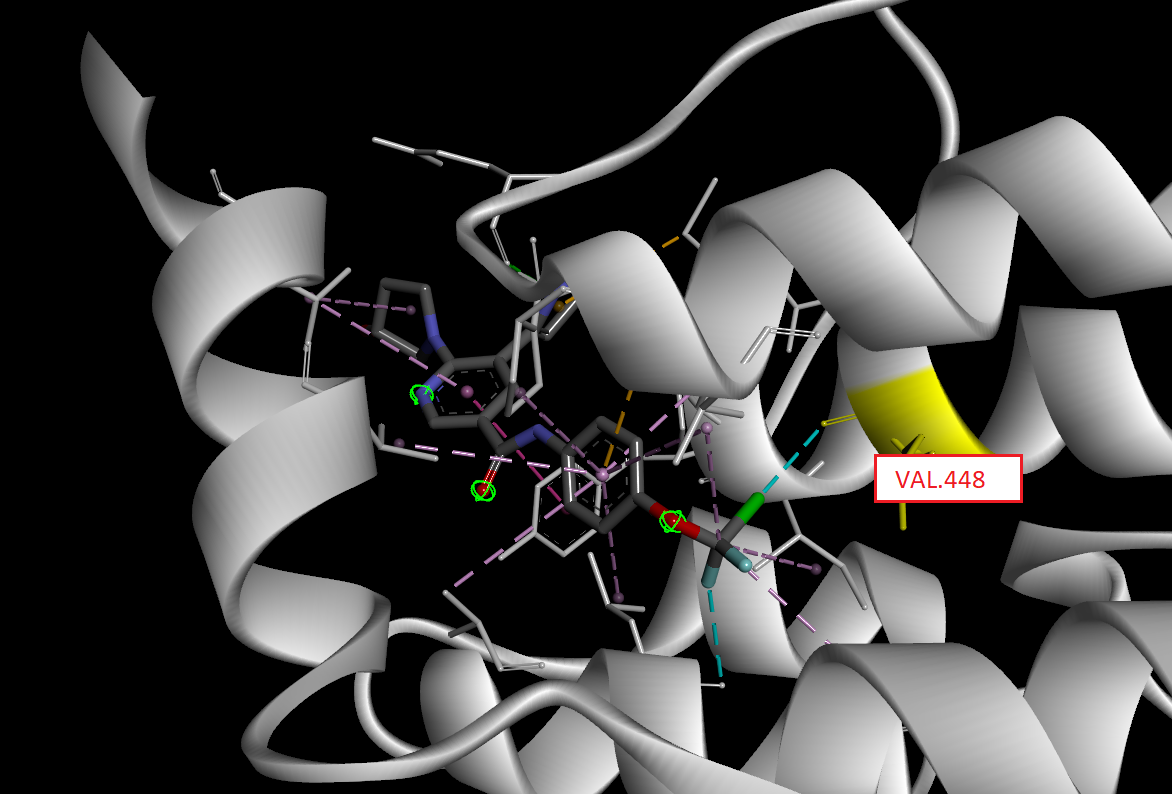}
\caption{Visualization of residue VAL.448 of protein ABL1-Asciminib (PDB
5MO4). The residue VAL.448 is the binding site of ABL1-Asciminib. Figure best viewed in color.\label{fig:ABL1_Asciminib}}
\end{figure}

We measure the frequency of mutation points in the ABL1 kinase domain over 500 counterfactual instances made from the Davis dataset (see Fig.~\ref{fig:prot_mutation_pos}). Top common mutation points are MET.244, LYS.247, VAL.260, GLU.286, LYS.291, and VAL.448. Importantly, most of these are, or in close proximity to, known binding sites of various drugs such as Nilotinib (LYS. 247, GLU.286, LYS.291, and MET.318, see Fig. \ref{fig:ABL1_Nilotinib}), Imatinib (LYS.247, GLU.286, and LYS.291), and Asciminib (VAL.448, see Fig. \ref{fig:ABL1_Asciminib}).

\subsection{ABL1-Nilotinib study case}

We choose Nilotinib for an in-depth case study because the ABL1-Nilotinib has the smallest prediction error in all ABL1-drugs pairs that also have the interaction and crystal structure available for assessment (c.f., PDB 3CS9).

\begin{figure}[h]
\centering{}\includegraphics[width=0.5\textwidth]{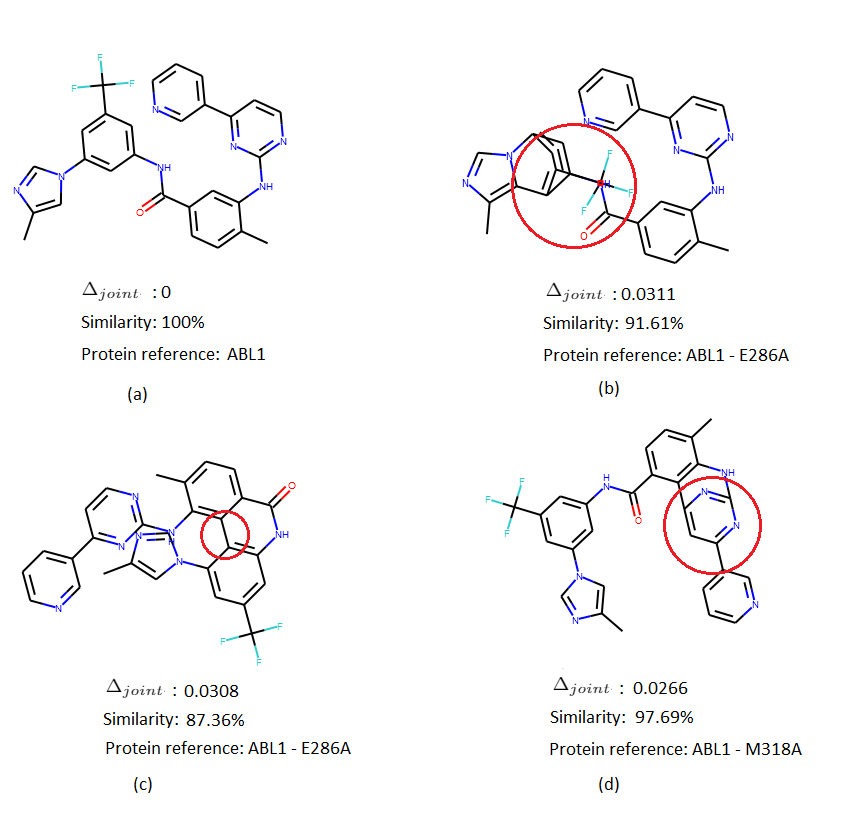}
\caption{Panel (a) is the original Nilotinib molecule. Panels (b)-(d) are three high $\Delta_{joint}$ counterfactual
instances of Nilotinib coupled with counterfactual instances of ABL1 as protein reference. The modification is circled in red. The counterfactual samples explain the interaction between the drug substructures and residues. For example in panel (b), the (trifluoromethyl)benzenes and residue E286 are modified which leads to the highest $\Delta_{joint}$. Therefore, MACDA suggests that the interaction between (trifluoromethyl)benzenes and E286 contributes most to the DTA model decision. \label{fig:Nilotinib_org_cf}}
 
\end{figure}

\begin{figure}[h]
\centering{}\includegraphics[width=0.45\textwidth]{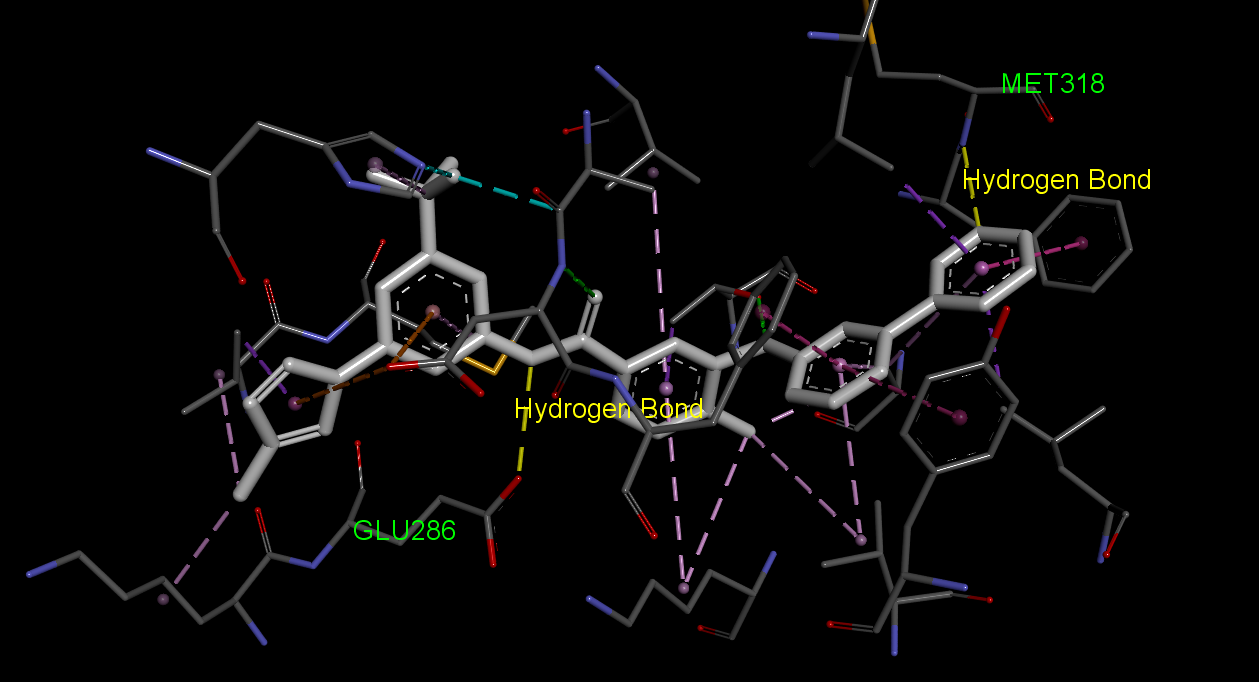}
\caption{The hydrogen bond between GLU.286 of ABL1 and the (trifluoromethyl)benzenes group
of Nilotinib (PDB 3CS9). Our model suggests cooperative binding between this position which seems plausible given the crystal structure. Figure best viewed in color. \label{fig:nilotinib_abl1}}
\end{figure}

Fig.~\ref{fig:Nilotinib_org_cf}
shows the 3 Nilotinib counterfactuals that have the highest $\Delta_{joint}$ when interacting with ABL1. 
In counterfactuals (b) and (c), the (trifluoromethyl)benzenes and residue GLU.286 are modified which leads to high $\Delta_{joint}$. We interpret this to mean that the interaction betwen the (trifluoromethyl)benzenes and E286 contributes strongly to the DTA model prediction for ABL1-Nilotinib. 
This also seems to be a biologically plausible binding mechanism.
Based on the crystal structure, there is a hydrogen bond between the (trifluoromethyl)benzenes and the GLU.286 (see Fig. \ref{fig:nilotinib_abl1}). 
In counterfactual (d), it can be suggested that the DTA model also takes into account the interaction between pyrimidine and MET.318 in its prediction. 

%% file: conclusion.tex
We have proposed a multi-agent reinforcement learning framework named
MACDA (Multi-Agent Counterfactual Drug-target binding Affinity) to
generate counterfactual explanations for the drug-target binding affinity
model. To address the discrete molecule graphs and protein sequences,
we use reinforcement learning to generate the counterfactual instances
which maximize the change in the binding affinity and the similarity
between counterfactual instances and original instances. To address
the two-input problem of drug-target binding affinity prediction model,
we use multi-agent reinforcement learning. Our multi-agent RL framework
consists of a protein agent and a drug molecule agent. Both agents
cooperate with each other to optimize the common multi-objective reward.
The experiments show that the proposed framework provides a stable
learning process and generates the counterfactual instances which
maximize the binding affinity change and minimize the joint change in the
input.

This work opens opportunities for future research. The counterfactual
instances for protein sequences are currently limited to the alanine scanning
process which is single point alanine mutation. Furthermore, the binding between drug and
target protein does not simply rely on a single residue but on a subset
of residues around the binding pocket. Instead of scanning using a
single residue, protein sequence motifs could be used. For the drug molecules, the counterfactual instances can
be generated at the fragment level instead of the atom level to maintain
the drug-likeness of the molecules. These all present open problems whose solutions could further improve joint affinity prediction and explanation parsimony. 

%% file: main.bbl
\begin{thebibliography}{}

\bibitem[Bickerton {\em et~al.}(2012)Bickerton, Paolini, Besnard, Muresan, and
  Hopkins]{Bickerton2012QuantifyingDrugs}
Bickerton, G.~R.  {\em et~al.} (2012).
\newblock {Quantifying the chemical beauty of drugs}.
\newblock {\em Nature Chemistry\/}, {\bf 4}(2), 90--98.

\bibitem[Cheng {\em et~al.}(2020)Cheng, Ming, and Qu]{Cheng2020DECE:Models}
Cheng, F.  {\em et~al.} (2020).
\newblock {DECE: Decision Explorer with Counterfactual Explanations for Machine
  Learning Models}.
\newblock {\em IEEE Transactions on Visualization and Computer Graphics\/}.

\bibitem[Davis {\em et~al.}(2011)Davis, Hunt, Herrgard, Ciceri, Wodicka,
  Pallares, Hocker, Treiber, and Zarrinkar]{Davis2011ComprehensiveSelectivity}
Davis, M.~I.  {\em et~al.} (2011).
\newblock {Comprehensive analysis of kinase inhibitor selectivity}.
\newblock {\em Nature Biotechnology\/}, {\bf 29}(11), 1046--1051.

\bibitem[Dhurandhar {\em et~al.}(2018)Dhurandhar, Chen, Luss, Tu, Ting,
  Shanmugam, and Das]{Dhurandhar2018ExplanationsNegatives}
Dhurandhar, A.  {\em et~al.} (2018).
\newblock {Explanations based on the Missing: Towards Contrastive Explanations
  with Pertinent Negatives}.
\newblock In {\em Advances in Neural Information Processing Systems\/}, pages
  592--603.

\bibitem[Fischer {\em et~al.}(2004)Fischer, Rovatsos, and
  Weiss]{Fischer2004HierarchicalCoordination}
Fischer, F.  {\em et~al.} (2004).
\newblock {Hierarchical Reinforcement Learning in Communication-Mediated
  Multiagent Coordination}.
\newblock In {\em Proceedings of the Third International Joint Conference on
  Autonomous Agents and Multiagent Systems-Volume 3\/}, pages 1334--1335.

\bibitem[Goyal {\em et~al.}(2019)Goyal, Wu, Ernst, Batra, Parikh, and
  Lee]{Goyal2019CounterfactualExplanations}
Goyal, Y.  {\em et~al.} (2019).
\newblock {Counterfactual Visual Explanations}.
\newblock In K.~Chaudhuri and R.~Salakhutdinov, editors, {\em Proceedings of
  the 36th International Conference on Machine Learning\/}, volume~97 of {\em
  Proceedings of Machine Learning Research\/}, pages 2376--2384. PMLR.

\bibitem[Gray {\em et~al.}(2017)Gray, Hause, and
  Fowler]{Gray2017AnalysisSubstitutions}
Gray, V.~E.  {\em et~al.} (2017).
\newblock {Analysis of Large-Scale Mutagenesis Data To Assess the Impact of
  Single Amino Acid Substitutions}.
\newblock {\em Genetics\/}, {\bf 207}(1), 53--61.

\bibitem[Hendricks {\em et~al.}(2016)Hendricks, Akata, Rohrbach, Donahue,
  Schiele, and Darrell]{Hendricks2016GeneratingExplanations}
Hendricks, L.~A.  {\em et~al.} (2016).
\newblock {Generating Visual Explanations}.
\newblock In {\em European Conference on Computer Vision\/}, pages 3--19.

\bibitem[Iqbal and Sha(2019)Iqbal and
  Sha]{Iqbal2019Actor-Attention-CriticLearning}
Iqbal, S. and Sha, F. (2019).
\newblock {Actor-Attention-Critic for Multi-Agent Reinforcement Learning}.
\newblock In {\em International Conference on Machine Learning\/}, pages
  2961--2970.

\bibitem[Jiang {\em et~al.}(2020)Jiang, Li, Zhang, Wang, Wang, Yuan, and
  Wei]{Jiang2020DrugtargetMaps}
Jiang, M.  {\em et~al.} (2020).
\newblock {Drug–target affinity prediction using graph neural network and
  contact maps}.
\newblock {\em RSC Advances\/}, {\bf 10}(35), 20701--20712.

\bibitem[Jorgensen {\em et~al.}(1983)Jorgensen, Chandrasekhar, Madura, Impey,
  and Klein]{Jorgensen1983ComparisonWater}
Jorgensen, W.~L.  {\em et~al.} (1983).
\newblock {Comparison of simple potential functions for simulating liquid
  water}.
\newblock {\em The Journal of Chemical Physics\/}, {\bf 79}(2), 926--935.

\bibitem[Konda and Borkar(1999)Konda and
  Borkar]{Konda1999Actor-CriticTypeProcesses}
Konda, V.~R. and Borkar, V.~S. (1999).
\newblock {Actor-Critic–Type Learning Algorithms for Markov Decision
  Processes}.
\newblock {\em SIAM Journal on control and Optimization\/}, {\bf 38}(1),
  94--123.

\bibitem[Li {\em et~al.}(2016)Li, Monroe, and
  Jurafsky]{Li2016UnderstandingErasure}
Li, J.  {\em et~al.} (2016).
\newblock {Understanding Neural Networks through Representation Erasure}.
\newblock {\em arXiv preprint arXiv:1612.08220\/}.

\bibitem[Littman(1994)Littman]{Littman1994MarkovLearning}
Littman, M.~L. (1994).
\newblock {Markov games as a framework for multi-agent reinforcement learning}.
\newblock In {\em Machine Learning Proceedings 1994\/}, pages 157--163.
  Elsevier.

\bibitem[McCloskey {\em et~al.}(2019)McCloskey, Taly, Monti, Brenner, and
  Colwell]{McCloskey2019UsingChemistry}
McCloskey, K.  {\em et~al.} (2019).
\newblock {Using attribution to decode binding mechanism in neural network
  models for chemistry}.
\newblock {\em Proceedings of the National Academy of Sciences\/}, {\bf
  116}(24), 11624--11629.

\bibitem[Meng {\em et~al.}(1992)Meng, Shoichet, and
  Kuntz]{Meng1992AutomatedEvaluation}
Meng, E.~C.  {\em et~al.} (1992).
\newblock {Automated docking with grid-based energy evaluation}.
\newblock {\em Journal of Computational Chemistry\/}, {\bf 13}(4), 505--524.

\bibitem[Morimura {\em et~al.}(2009)Morimura, Uchibe, Yoshimoto, and
  Doya]{Morimura2009AAlgorithm}
Morimura, T.  {\em et~al.} (2009).
\newblock {A Generalized Natural Actor-Critic Algorithm}.
\newblock {\em Advances in Neural Information Processing Systems\/}, {\bf 22},
  1312--1320.

\bibitem[Mothilal {\em et~al.}(2020)Mothilal, Sharma, and
  Tan]{Mothilal2020ExplainingExplanations}
Mothilal, R.~K.  {\em et~al.} (2020).
\newblock {Explaining Machine Learning Classifiers through Diverse
  Counterfactual Explanations}.
\newblock In {\em Proceedings of the 2020 Conference on Fairness,
  Accountability, and Transparency\/}, pages 607--617.

\bibitem[Nguyen {\em et~al.}(2020)Nguyen, Le, Quinn, Nguyen, Le, and
  Venkatesh]{Nguyen2020GraphDTA:Networks}
Nguyen, T.  {\em et~al.} (2020).
\newblock {GraphDTA: Predicting drug-target binding affinity with graph neural
  networks}.
\newblock {\em Bioinformatics\/}.

\bibitem[Numeroso and Bacciu(2020)Numeroso and
  Bacciu]{Numeroso2020ExplainingCounterfactuals}
Numeroso, D. and Bacciu, D. (2020).
\newblock {Explaining Deep Graph Networks with Molecular Counterfactuals}.
\newblock {\em Advances in Neural Information Processing Systems, Workshop on
  Machine Learning for Molecules\/}.

\bibitem[Numeroso and Bacciu(2021)Numeroso and
  Bacciu]{Numeroso2021MEG:Networks}
Numeroso, D. and Bacciu, D. (2021).
\newblock {MEG: Generating Molecular Counterfactual Explanations for Deep Graph
  Networks}.
\newblock In {\em Proceedings of International Joint Conference on Neural
  Networks\/}.

\bibitem[{\"{O}}zt{\"{u}}rk {\em et~al.}(2018){\"{O}}zt{\"{u}}rk,
  {\"{O}}zg{\"{u}}r, and Ozkirimli]{Ozturk2018DeepDTA:Prediction}
{\"{O}}zt{\"{u}}rk, H.  {\em et~al.} (2018).
\newblock {DeepDTA: deep drug–target binding affinity prediction}.
\newblock {\em Bioinformatics\/}, {\bf 34}(17), i821–i829.

\bibitem[{\"{O}}zt{\"{u}}rk {\em et~al.}(2019){\"{O}}zt{\"{u}}rk, Ozkirimli,
  and {\"{O}}zg{\"{u}}r]{Ozturk2019WideDTA:Affinity}
{\"{O}}zt{\"{u}}rk, H.  {\em et~al.} (2019).
\newblock {WideDTA: prediction of drug-target binding affinity}.
\newblock {\em arXiv preprint arXiv:1902.04166\/}.

\bibitem[Pearl {\em et~al.}(2016)Pearl, Glymour, and
  Jewell]{Pearl2016CausalPrimer}
Pearl, J.  {\em et~al.} (2016).
\newblock {\em {Causal Inference in Statistics: A Primer}\/}.
\newblock John Wiley {\&} Sons.

\bibitem[Perez-Liebana {\em et~al.}(2019)Perez-Liebana, Hofmann, Mohanty, Kuno,
  Kramer, Devlin, Gaina, and Ionita]{Perez-Liebana2019TheCompetition}
Perez-Liebana, D.  {\em et~al.} (2019).
\newblock {The Multi-Agent Reinforcement Learning in MalmO (MARLO)
  Competition}.
\newblock {\em arXiv preprint arXiv:1901.08129\/}.

\bibitem[Pope {\em et~al.}(2019)Pope, Kolouri, Rostami, Martin, and
  Hoffmann]{Pope2019ExplainabilityNetworks}
Pope, P.~E.  {\em et~al.} (2019).
\newblock {Explainability Methods for Graph Convolutional Neural Networks}.
\newblock In {\em "Proceedings of the IEEE Conference on Computer Vision and
  Pattern Recognition"\/}, pages 10772--10781.

\bibitem[Preuer {\em et~al.}(2019)Preuer, Klambauer, Rippmann, Hochreiter, and
  Unterthiner]{Preuer2019InterpretableDiscovery}
Preuer, K.  {\em et~al.} (2019).
\newblock {Interpretable Deep Learning in Drug Discovery}.
\newblock In {\em Explainable AI: Interpreting, Explaining and Visualizing Deep
  Learning\/}, pages 331--345. Springer.

\bibitem[Pullman(2013)Pullman]{Pullman2013IntermolecularForces}
Pullman, A. (2013).
\newblock {\em {Intermolecular Forces}\/}, volume~14.
\newblock Springer Science {\&} Business Media.

\bibitem[Raha {\em et~al.}(2007)Raha, Peters, Wang, Yu, Wollacott, Westerhoff,
  and Merz~Jr]{Raha2007TheDesign}
Raha, K.  {\em et~al.} (2007).
\newblock {The role of quantum mechanics in structure-based drug design}.
\newblock {\em Drug Discovery Today\/}, {\bf 12}(17-18), 725--731.

\bibitem[Rodr{\'{i}}guez-P{\'{e}}rez and
  Bajorath(2020)Rodr{\'{i}}guez-P{\'{e}}rez and
  Bajorath]{Rodriguez-Perez2020InterpretationValues}
Rodr{\'{i}}guez-P{\'{e}}rez, R. and Bajorath, J. (2020).
\newblock {Interpretation of compound activity predictions from complex machine
  learning models using local approximations and shapley values}.
\newblock {\em Journal of Medicinal Chemistry\/}, {\bf 63}(16), 8761--8777.

\bibitem[Ryu {\em et~al.}(2018)Ryu, Lim, Hong, and Kim]{Ryu2018DeeplyNetwork}
Ryu, S.  {\em et~al.} (2018).
\newblock {Deeply learning molecular structure-property relationships using
  attention-and gate-augmented graph convolutional network}.
\newblock {\em arXiv preprint arXiv:1805.10988\/}.

\bibitem[Shang {\em et~al.}(2018)Shang, Liu, Chen, Sun, Lu, Yi, and
  Bi]{Shang2018EdgeNetworksb}
Shang, C.  {\em et~al.} (2018).
\newblock {Edge Attention-based Multi-Relational Graph Convolutional Networks}.

\bibitem[Sutton and Barto(2018)Sutton and
  Barto]{Sutton2018ReinforcementIntroduction}
Sutton, R.~S. and Barto, A.~G. (2018).
\newblock {\em {Reinforcement learning: An introduction}\/}.
\newblock MIT press.

\bibitem[Tan(1993)Tan]{Tan1993Multi-AgentAgents}
Tan, M. (1993).
\newblock {Multi-Agent Reinforcement Learning: Independent vs. Cooperative
  Agents}.
\newblock In {\em Proceedings of the Tenth International Conference on Machine
  Learning\/}, pages 330--337.

\bibitem[Thafar {\em et~al.}(2019)Thafar, Raies, Albaradei, Essack, and
  Bajic]{Thafar2019ComparisonAffinities}
Thafar, M.  {\em et~al.} (2019).
\newblock {Comparison Study of Computational Prediction Tools for Drug-Target
  Binding Affinities}.
\newblock {\em Frontiers in Chemistry\/}, {\bf 7}.

\bibitem[Torng and Altman(2019)Torng and Altman]{Torng2019GraphInteractions}
Torng, W. and Altman, R.~B. (2019).
\newblock {Graph Convolutional Neural Networks for Predicting Drug-Target
  Interactions}.
\newblock {\em Journal of Chemical Information and Modeling\/}, {\bf 59}(10),
  4131--4149.

\bibitem[Tri {\em et~al.}(2020)Tri, Nguyen, Le, and
  Tran]{Tri2020GEFA:Prediction}
Tri, N.  {\em et~al.} (2020).
\newblock {GEFA: Early Fusion Approach in Drug-Target Affinity Prediction}.
\newblock {\em Advances in Neural Information Processing Systems, Workshop on
  Machine Learning for Structural Biology\/}.

\bibitem[Tsang {\em et~al.}(2018)Tsang, Liu, Purushotham, Murali, and
  Liu]{tsang_neural_2018}
Tsang, M.  {\em et~al.} (2018).
\newblock Neural {Interaction} {Transparency} ({NIT}): {Disentangling}
  {Learned} {Interactions} for {Improved} {Interpretability}.
\newblock In S.~Bengio, H.~Wallach, H.~Larochelle, K.~Grauman, N.~Cesa-Bianchi,
  and R.~Garnett, editors, {\em Advances in {Neural} {Information} {Processing}
  {Systems} 31\/}, pages 5804--5813. Curran Associates, Inc.

\bibitem[Vaswani {\em et~al.}(2017)Vaswani, Shazeer, Parmar, Uszkoreit, Jones,
  Gomez, Kaiser, and Polosukhin]{Vaswani2017AttentionNeed}
Vaswani, A.  {\em et~al.} (2017).
\newblock {Attention is all you need}.
\newblock In {\em Advances in Neural Information Processing Systems\/}, pages
  5998--6008.

\bibitem[Veli{\v{c}}kovi{\'{c}} {\em et~al.}(2018)Veli{\v{c}}kovi{\'{c}},
  Cucurull, Casanova, Romero, Li{\`{o}}, and
  Bengio]{Velickovic2018GraphNetworks}
Veli{\v{c}}kovi{\'{c}}, P.  {\em et~al.} (2018).
\newblock {Graph Attention Networks}.
\newblock {\em International Conference on Learning Representations\/}.

\bibitem[Vermeire and Martens(2020)Vermeire and
  Martens]{Vermeire2020ExplainableCounterfactual}
Vermeire, T. and Martens, D. (2020).
\newblock {Explainable Image Classification with Evidence Counterfactual}.
\newblock {\em arXiv preprint arXiv:2004.07511\/}.

\bibitem[Wachter {\em et~al.}(2017)Wachter, Mittelstadt, and
  Russell]{Wachter2017CounterfactualGDPR}
Wachter, S.  {\em et~al.} (2017).
\newblock {Counterfactual explanations without opening the black box: Automated
  decisions and the GDPR}.
\newblock {\em Harvard Journal of Law {\&} Technology\/}, {\bf 31}, 841.

\bibitem[Ying {\em et~al.}(2019)Ying, Bourgeois, You, Zitnik, and
  Leskovec]{Ying2019GNNExplainer:Networks}
Ying, R.  {\em et~al.} (2019).
\newblock {GNNExplainer: Generating Explanations for Graph Neural Networks}.
\newblock {\em Advances in Neural Information Processing Systems\/}, {\bf 32},
  9240.

\bibitem[Zheng {\em et~al.}(2020)Zheng, Li, Chen, Xu, and
  Yang]{Zheng2020PredictingSystem}
Zheng, S.  {\em et~al.} (2020).
\newblock {Predicting drug–protein interaction using quasi-visual question
  answering system}.
\newblock {\em Nature Machine Intelligence\/}, {\bf 2}(2), 134--140.

\bibitem[Zhou and Troyanskaya(2015)Zhou and Troyanskaya]{zhou_predicting_2015}
Zhou, J. and Troyanskaya, O.~G. (2015).
\newblock Predicting effects of noncoding variants with deep learning–based
  sequence model.
\newblock {\em Nature Methods\/}, {\bf 12}(10), 931--934.

\bibitem[Zhou {\em et~al.}(2019)Zhou, Kearnes, Li, Zare, and
  Riley]{Zhou2019OptimizationLearning}
Zhou, Z.  {\em et~al.} (2019).
\newblock {Optimization of Molecules via Deep Reinforcement Learning}.
\newblock {\em Scientific Reports\/}, {\bf 9}(1), 1--10.

\end{thebibliography}
